\documentclass[runningheads]{llncs}

\usepackage{eccv}

\usepackage{eccvabbrv}

\usepackage{graphicx}
\usepackage{booktabs}
\usepackage{colortbl}
\usepackage{multirow}
\usepackage{float}
\usepackage{wasysym}
\DeclareRobustCommand{\rev}[1]{#1}
\usepackage[accsupp]{axessibility}  % Improves PDF readability for those with disabilities.
\usepackage{hyperref}
\hypersetup{
    hidelinks,
    pdfborder={0 0 1},
    citecolor=green
}
\begin{document}

% ---------------------------------------------------------------
\title{EgoSafe: A First-Person Mobile-Captured Benchmark for Visual Safety Understanding}

\titlerunning{EgoSafe}

% Author metadata is kept in a separate file for the arXiv package.
\newcommand{\arxivauthorlist}{%
Yuyun Chen\inst{1}\thanks{These authors contributed equally to this work.} \and
Tianao Li\inst{1}\ensuremath{^\star} \and
TianQuan Feng\inst{1} \and
Cen Chen\inst{1} \and
Huiping Zhuang\inst{1} \and
Hao Peng\inst{2} \and
Ziqian Zeng\inst{1}\thanks{Corresponding author.}}

\newcommand{\arxivauthorrunning}{EgoSafe}

\newcommand{\arxivinstitute}{%
South China University of Technology\\
\email{202430300212@mail.scut.edu.cn, 202464870164@mail.scut.edu.cn}\\
\email{202364820431@mail.scut.edu.cn, chencen@scut.edu.cn}\\
\email{hpzhuang@scut.edu.cn, zqzeng@scut.edu.cn}
\and
Beihang University\\
\email{penghao@buaa.edu.cn}} 

\author{\arxivauthorlist}
\authorrunning{\arxivauthorrunning}
\institute{\arxivinstitute}

\maketitle

% -----------------------------------------------------------------------------
% Abstract
% -----------------------------------------------------------------------------
\begin{abstract}
Reliable visual safety understanding in real-world scenarios demands more than just pattern matching, and requires \textbf{evidence-based reasoning under limited visibility}. While Large Vision-Language Models (LVLMs) demonstrate impressive semantic alignment on standard benchmarks, they often struggle to distinguish between superficial correlation and evidence-based reasoning when grounded in the dynamic, partially observable nature of first-person experiences. Existing evaluations, dominated by third-person surveillance footage and binary classification metrics, fail to expose this cognitive gap. To address this, we introduce \textbf{EgoSafe-Bench}, a benchmark specifically designed to evaluate models' reasoning ability in egocentric safety scenarios. It comprises \textbf{12,000 unique evaluation samples}, generated by pairing each of the \textbf{3,000 video clips} with a QA chain governed by our proposed \textbf{Hierarchical Reasoning Evaluation (HRE)} framework. Unlike standard benchmarks, \textbf{HRE} mandates a rigorous reasoning trajectory from initial feature anchoring to blind-spot deduction and intent inference, thereby enforcing reasoning consistency and penalizing shortcut-based predictions.
Extensive evaluations of state-of-the-art LVLMs (e.g., Qwen3-VL~\cite{qwen3technicalreport}, Gemini~\cite{team2023gemini,team2024gemini}, VideoLLaMA 3~\cite{damonlpsg2025videollama3}) reveal a significant perception-reasoning decoupling, whereby models often achieve high descriptive scores but exhibit notable fragility in reasoning based on contextual evidence. Our work provides both a challenging dataset and a systematic evaluation framework to foster the development of logically robust video understanding systems.

\keywords{Video Understanding \and Violence detection \and Benchmark Construction}
\end{abstract}

% -----------------------------------------------------------------------------
% Introduction
% -----------------------------------------------------------------------------
\section{Introduction}
\label{sec:intro}

% [PLACEHOLDER FOR TEASER FIGURE]
% 建议在这里放一张 "Teaser Figure"。
% 左边展示：传统 Benchmark (CCTV视角，清晰，任务是简单的"Fight/No Fight")。
% 右边展示：EgoSafe (第一人称视角，模糊/遮挡，任务是"推理盲区里的动作" -> "判断意图")。
% Caption 核心：Contrast between traditional "God-view" recognition and egocentric forensic reasoning.
\begin{figure}[t]
  \centering
  \includegraphics[width=\linewidth]{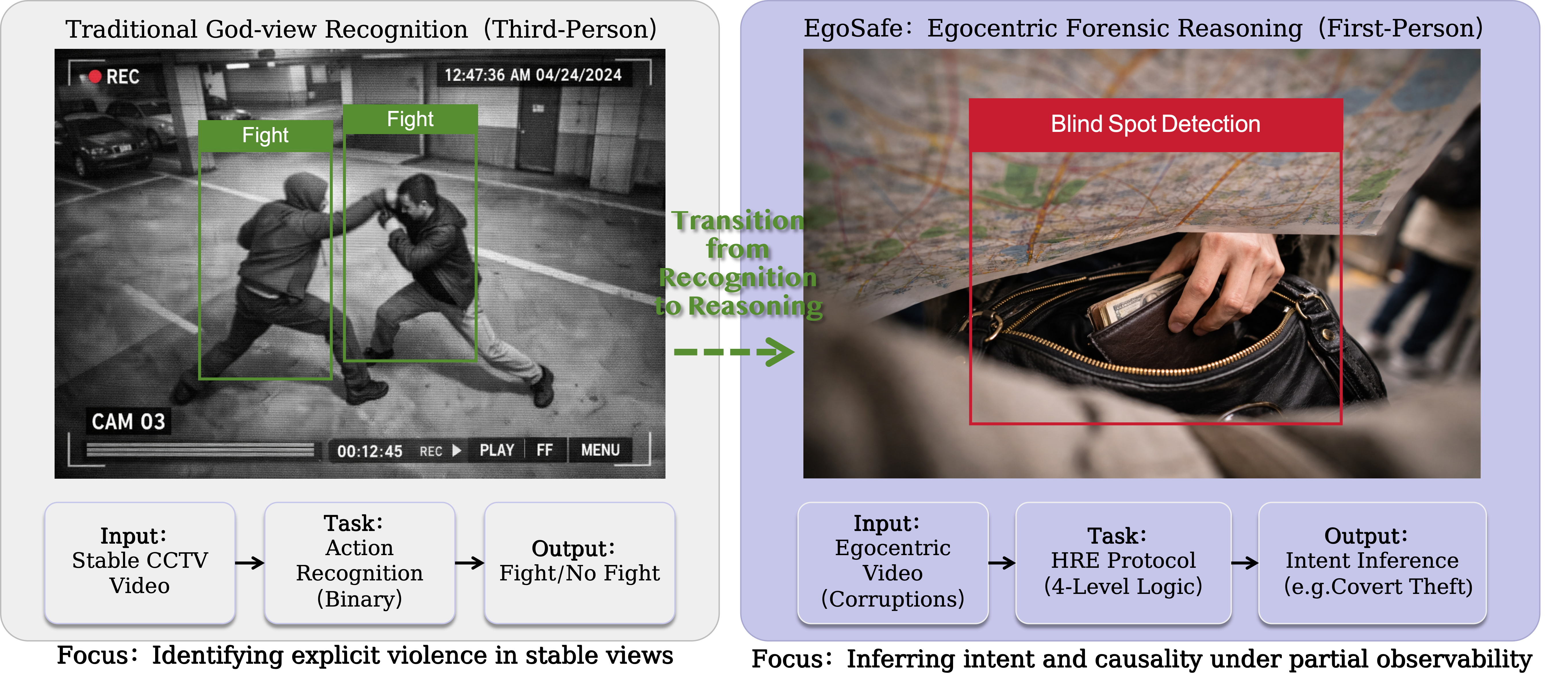}
  % \fbox{\rule{0pt}{3in} \rule{0.9\linewidth}{0pt}} % 占位符
  \caption{\textbf{From Recognition to Evidence-Based Reasoning.} While traditional benchmarks focus on identifying explicit violence in stable third-person views (Left), real-world safety understanding requires inferring intent and causality under partial observability (Right). \textbf{EgoSafe-Bench} challenges LVLMs to perform hierarchical reasoning, transitioning from anchoring visual evidence to deducing covert actions in blind spots.}
  \label{fig:teaser}
\end{figure}

Visual safety understanding is a cornerstone of autonomous agents and social governance~\cite{sabha2024towards,pujol2020soft}.For autonomous agents, it dictates the reliability of their interactions with the physical world, directly impacting human safety and operational success. Concurrently, in social governance, guaranteeing the integrity and unbiased interpretation of visual data is critical for formulating equitable policies, ensuring public security, and maintaining trust in automated systems. Unlike generic video understanding, safety-critical analysis in the real world is inherently a problem of \textbf{evidence-based reasoning under limited visibility}. In practical deployments, whether via body-worn cameras or mobile devices, critical visual evidence is frequently fragmented, occluded, or temporally implicit. Reliable judgment, therefore, depends not only on perceiving \textit{what} is visible but also on logically inferring \textit{how} an event progresses, particularly when key interactions occur within visual blind spots or involve subtle behavioral escalation.

The rapid evolution of Large Vision-Language Models (LVLMs), from early architectures~\cite{damonlpsg2023videollama,lin2023video} to recent frontiers like Qwen3-VL~\cite{qwen3technicalreport}, VideoLLaMA 3~\cite{damonlpsg2025videollama3}, and InternVL 3.5~\cite{wang2025internvl3_5} has demonstrated exceptional cross-modal capabilities. Concurrently, safety benchmarks have also progressed from binary surveillance classification (e.g., UCF-Crime~\cite{Sultani_2018_CVPR}, RWF-2000~\cite{9412502}) to more complex multimodal~\cite{Wu2020not}, fine-grained~\cite{xiang2024video}, and diverse~\cite{kollias2025dvd} evaluations. While egocentric or casual reasoning datasets focusing on daily life have seen significant development, there remains a lack of datasets specifically designed for violent situations.~\cite{Grauman_2022_CVPR, xiao2021next, li2022from, wang2024egovid, zhu2023egoobjects, 9084270, grauman2024ego}

However, there exist two fundamental gaps that prevent existing benchmarks from revealing the true reasoning capabilities of LVLMs in safety scenarios:

\textbf{1. The Reasoning Gap: From Superficial Visual Cues to Evidence-based Reasoning.}
Current evaluation metrics often combine \textit{pattern matching capability} with true \textit{reasoning capability}. Benchmarks such as Video Violence Rating~\cite{xiang2024video} or MVBench~\cite{Li_2024_CVPR} typically evaluate models using isolated queries or coarse global scores. Consequently, these benchmarks practically only validate the models' superficial visual recognition rather than effectively evaluating their deep reasoning capabilities. For instance, a model can achieve a high score merely by detecting a gun, without needing to comprehend whether the broader context implies a real threat, a training exercise, or a defensive measure. Recent research has revealed that LVLMs achieve exceptionally high recall on this task.\cite{ditchfield2025comparative} This performance profile indicates that models heavily rely on superficial visual cues (e.g., rapid motion or weapon-like objects) to determine danger, rather than exhibiting real evidence-based reasoning ability. Unless we explicitly evaluate how a model maintains reasoning consistency as visual evidence evolves over time, we cannot reliably trust its judgments in safety-critical applications.

\textbf{2. The Perspective Gap: From God-View to Egocentric Vision.}
Existing safety benchmarks, such as UCF-Crime~\cite{Sultani_2018_CVPR}, RWF-2000~\cite{9412502}, and XD-Violence~\cite{Wu2020not}, predominantly rely on ``studio-quality'' third-person footage or static surveillance streams. These datasets possess stable viewpoints and clear visual evidence, treating safety understanding as a task that benefits from a god view capable of capturing the entire trajectory of a violent event. In contrast, real-world safety scenarios are inherently first-person (egocentric). In practice, while third-person ``god-view'' videos are frequently utilized for post-hoc analysis, proactive harm prevention requires alerting the victim the moment a danger emerges. This necessitates models that are highly capable of reasoning within real egocentric scenarios. Real-world scenarios eliminate the global comprehensiveness of a ``god-view'', creating an environment where visual evidence is fragmented and key interactions frequently occur within \textbf{visual blind spots}. Additionally, egocentric vision introduces limited visibility like camera shake and motion blur, which severely degrade standard pattern matching. Confronted with this highly fragmented visual input, simply identifying surface-level visual features is no longer adequate. However, models that rely on pattern matching on stable surveillance streams lack the reasoning required to deduce events within visual blind spots. Consequently, they inherently struggle to generalize to high-variance, real-world scenarios where the observer is deeply involved in the scene.

% Our Solution / Contributions
To bridge these gaps, we propose \textbf{EgoSafe-Bench}, a pioneering benchmark tailored for first-person visual safety understanding. Unlike previous efforts, we shift the focus from ``what is happening'' to ``why it is unsafe'', enforcing a rigorous check on the model's cognitive process. Our main contributions are:

First, to address the lack of safety data from an egocentric perspective, we present a high-fidelity dataset of \textbf{3,000} self-captured egocentric video clips and \textbf{12,000} hierarchical evaluation samples(\textbf{EgoSafe dataset}). The data intentionally incorporates authentic environmental noise (e.g., lighting shifts, jitter) and fragments in ego view to simulate real-world perceptual challenges.
    
Second, to evaluate whether LVLMs are capable of evidence-based reasoning rather than mere superficial pattern matching. We introduce a novel evaluation framework that structures QA chains through an explicit four-tier \textbf{Hierarchical Reasoning Evaluation (HRE)}: (1) \textit{Initial Feature Anchoring}, (2) \textit{Dynamic Blind Spot Construction}, (3) \textit{Covert Action Deduction}, and (4) \textit{Causal Result Qualitative Analysis}. This framework requires models to explicitly reason about invisible interactions and maintain reasoning consistency, rather than answering isolated questions by simply pattern matching.
    
Third,we conduct extensive evaluations of leading models, including Qwen3-VL~\cite{qwen3technicalreport} ,VideoLLaMA 3~\cite{damonlpsg2025videollama3}, and so on. Our experiments reveal a ``reasoning decoupling'' phenomenon, whereby models perform well on surface-level semantic description, their performance degrades sharply when forced to perform multi-step causal deduction, highlighting the necessity of logic-driven benchmarks.

\section{Related Work}
\label{sec:related_work}

\textbf{Large Vision-Language Models for Video.} 
LVLMs have rapidly evolved from basic image-text alignment to complex spatial-temporal reasoning. Early models like Video-LLaMA~\cite{damonlpsg2023videollama} and Video-LLaVA~\cite{lin2023video} pioneered visual-LLM bridging. Recently, frontier models including the Qwen-VL series~\cite{Qwen-VL, Qwen2VL, Qwen2.5-VL, qwen3technicalreport}, VideoLLaMA 2/3~\cite{damonlpsg2024videollama2, damonlpsg2025videollama3}, GPT-4~\cite{achiam2023gpt}, and Gemini~\cite{team2023gemini, team2024gemini}, alongside efficient architectures like MiniCPM-V~\cite{yu2025minicpmv45cookingefficient} and InternVL 3.5~\cite{wang2025internvl3_5}, have achieved state-of-the-art performance on generic benchmarks (e.g., MVBench~\cite{Li_2024_CVPR}). However, their capabilities often degrade in safety-critical scenarios characterized by egocentric motion and low-quality mobile captures. Our work specifically benchmarks these models under authentic environmental noise to target this gap.

\textbf{Violence Detection Benchmarks.} 
Traditional violence detection benchmarks, such as UCF-Crime~\cite{Sultani_2018_CVPR}, RWF-2000~\cite{9412502}, and XD-Violence~\cite{Wu2020not}, primarily formulate the problem as anomaly or binary classification using surveillance footage. Recent efforts like VVR~\cite{xiang2024video} and DVD~\cite{kollias2025dvd} introduce fine-grained intensity ratings and diverse real-world scenarios~\cite{kollias2025dvd}. Despite these advances, existing datasets predominantly rely on stable, third-person perspectives (e.g., CCTV), lacking the erratic camera dynamics and occlusions inherent in mobile recordings. Furthermore, their closed-domain evaluation metrics fail to probe a model's reasoning consistency. EgoSafe-Bench addresses this by introducing a first-person dataset coupled with a hierarchical QA framework to mandate evidence-based reasoning over superficial pattern matching.

\textbf{Egocentric and Causal Reasoning Benchmarks.} 
Beyond safety scenarios, egocentric video understanding has driven massive data collection efforts focusing on daily life and human-object interactions. Datasets like Ego4D~\cite{Grauman_2022_CVPR}, EPIC-KITCHENS~\cite{9084270}, and Ego-Exo4D~\cite{grauman2024ego} provide extensive first-person footage for action recognition, forecasting, and skilled activity analysis. Concurrently, causal and temporal reasoning in video has been explored through benchmarks like NExT-QA~\cite{xiao2021next} and representation studies~\cite{li2022from}, which focus on explaining the causality behind actions in general contexts. However, these datasets are heavily skewed towards benign, cooperative daily activities (e.g., cooking, crafting, fine-grained object manipulation~\cite{zhu2023egoobjects}). They fundamentally lack the adversarial dynamics, aggressive occlusions, and evidence-based reasoning under limited visibility that are crucial for real-world safety understanding.

\section{The EgoSafe-Bench Dataset}
\label{sec:dataset}

\subsection{Data Collection and Processing}
\textbf{Overview of the EgoSafe-Bench Dataset.} The EgoSafe-Bench dataset consists entirely of \textbf{first-person} videos captured by mobile devices, targeting the unpredictable and highly dynamic nature of real-world safety encounters. To support deep logical reasoning, we utilize high-fidelity \textbf{4K resolution}, ensuring that critical microscopic cues, such as subtle gestures and hidden objects, are well-preserved despite the unstable camera motion. \rev{All 250 source videos were staged by our team using an iPhone 14 Pro following pre-written scripts.}

A defining characteristic of EgoSafe-Bench is its emphasis on \textbf{complex visual noise}. We specifically construct the dataset in \textbf{highly challenging environments}, such as narrow corridors, dimly lit pathways and building entrances, featuring severe background clutter and dynamic light pollution. By naturally including these stressors, our dataset moves beyond the controlled settings of prior egocentric benchmarks. This provides a rigorous test for evaluating a model's reasoning under strong interference. Furthermore, EgoSafe-Bench covers a \textbf{wide range of security events}, from covert thefts to physical altercations. This variety ensures our dataset realistically reflects real-world, first-person risk scenarios, making it a reliable tool for assessing hazard recognition.

\textbf{Robustness Evaluation and Degradation Processing Methods.} While our raw footage inherently captures natural visual complexities, real-world safety crises often push these interferences to extreme limits. Because natural noise is inherently uncontrollable and unevenly distributed across videos, such noise is insufficient for a standardized stress test. To systematically quantify the models' reliability and identify their breaking points under worst-case scenarios, we draw inspiration from standard robustness benchmarks~\cite{DBLP:conf/iclr/HendrycksD19, DBLP:journals/corr/abs-2110-06513} and introduce a comprehensive evaluation featuring \textbf{11 visual degradations}.

Specifically, by \textbf{downsampling the original 4K footage to 1080p and 720p}, we simulate the compression artifacts commonly found in live-streamed body camera footage or older hardware, testing the model's performance under resolution degradation. Next, we \textbf{generate six variants of lighting conditions}, including increased and decreased brightness, to evaluate the model's performance in extreme lighting environments such as overexposure (e.g., strong artificial lights in a bar) or underexposure (e.g., dark alleys). Additionally, to simulate the severe motion blur that occurs during physical struggles, we combine \textbf{horizontal flipping, camera shake, and Gaussian blur}, thereby reproducing the extreme visual instability of real-world conflicts. Through the comprehensive testing of these degradation types, we evaluate not only the model's visual understanding capability but also its robustness across varying video qualities. This systematic degradation testing allows us to measure how effectively the model can maintain its reasoning ability when visual quality deteriorates, thereby providing a more comprehensive evaluation of its reliability in complex real-world environments.

\textbf{Ethical Compliance and Privacy Protection.} We strictly follow ethical guidelines regarding biometric data privacy. \rev{All scenarios were staged by our team, and all participants gave informed consent for academic use.} We apply solid masks to all identifiable faces and objects that might reveal sensitive personal information (such as ID cards and license plates) to \textbf{ensure irreversible data de-identification}. \rev{All processed data undergoes \textbf{a secondary manual review} to verify the effectiveness of the masking, ensuring that the de-identified dataset can be safely and compliantly released for public academic research.} Furthermore, these strict masking measures introduce an additional challenge for the models, requiring them to infer intent from body language and environmental context rather than relying on facial expressions.

\subsection{Hierarchical Reasoning Evaluation (HRE) Annotation}
To systematically evaluate the models' ability to handle the above uncertainties, we propose the \textbf{Hierarchical Reasoning Evaluation (HRE)} protocol. Traditional datasets often rely on simple pattern matching, allowing models to take ``shortcuts'' and guess answers using superficial visual cues~\cite{xiao2021next, li2022from, Li_2024_CVPR}. In contrast, \textbf{HRE} structures data annotation as a progressive \textbf{four-stage QA chain}, forcing the model to execute a structured reasoning process. We leverage the long-context reasoning capabilities of \textbf{Gemini-3-Pro}~\cite{team2023gemini,team2024gemini} to instantiate the QA chains.

The \textit{first stage focuses on visual anchoring}, requiring the model to identify key objects and initial states before any occlusion occurs. The \textit{second stage targets blind-spot recognition}, testing whether the model can perceive how critical visual information is artificially occluded. Building upon this, the \textit{third stage demands covert action deduction}, where the model must infer hidden interactions within the blind spot using surrounding contextual clues. The \textit{fourth stage involves causal outcome inference}, challenging the model to accurately determine the final nature of the event. This continuous QA chain ensures that the model must understand the entire process rather than relying on shortcut predictions.

\begin{figure*}
  \centering
  \includegraphics[width=\textwidth]{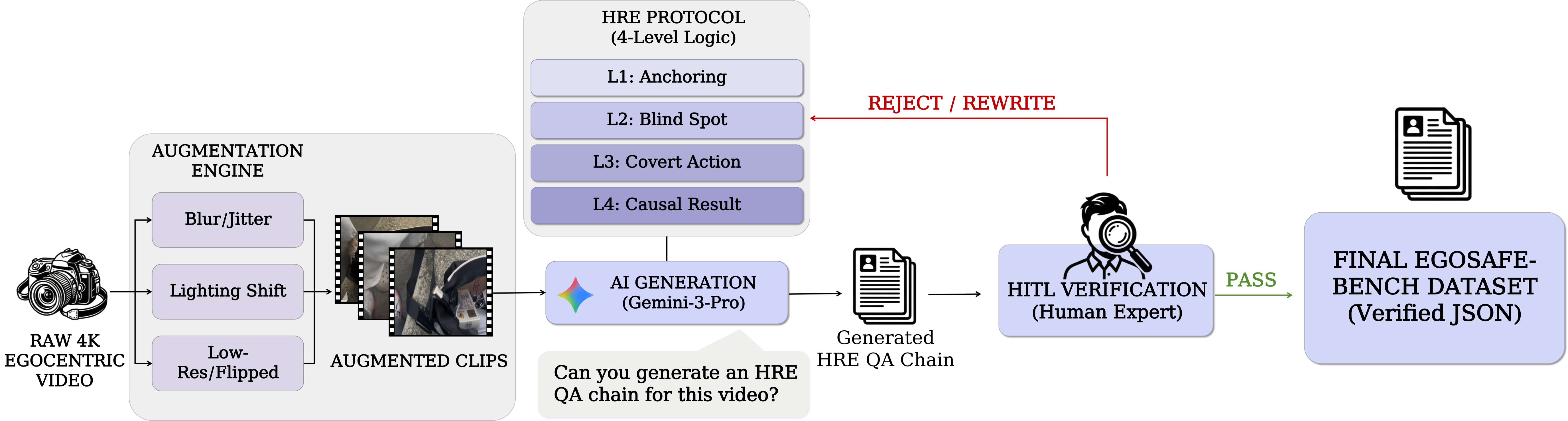}
  \caption{\textbf{Overview of the data generation and verification pipeline for EgoSafe-Bench.} A \textbf{human-in-the-loop (HITL)} verification mechanism ensures reasoning consistency, where flawed annotations are rejected and rewritten before being integrated into the final dataset.}
  \label{fig:pipeline_overview}
\end{figure*}

Furthermore, we introduce a \textbf{``human-in-the-loop''} verification mechanism into the annotation pipeline~\cite{maaz-etal-2024-video}. \rev{All machine-generated annotations are reviewed by independent verifiers.} \rev{They check for hallucinated content and assess whether the reasoning process is sound.} \rev{We retain only clips for which the verifiers agree on the event-type classification.} \rev{Any question-answering chain that fails to meet the factual-correctness or reasoning-consistency standard is rejected and rewritten.} This rigorous filtering ensures that EgoSafe-Bench serves as a reliable benchmark for evaluating machine reasoning.

\subsection{Dataset Statistics and Comparison}

Overall, EgoSafe-Bench establishes a dedicated benchmark for reasoning and robustness in critical safety scenarios. To systematically evaluate the visual robustness of models~\cite{DBLP:conf/iclr/HendrycksD19}, we start with 250 unique, high-fidelity source videos. \rev{Through a rigorous augmentation pipeline (detailed in Section 3.1), we retain each clean source and generate \textbf{11 variants for each video}, ultimately constructing a large-scale dataset comprising \textbf{3,000 clips}.} With an average duration of approximately 10 seconds per clip and a total of 8.3 hours of footage, the dataset features exceptionally high frame-level precision and superior visual resolution. Following the \textbf{HRE protocol}, each video is paired with a \textbf{4-turn hierarchical question-answering (QA) chain} that strictly corresponds to four reasoning levels, \textbf{anchoring, blind spot, covert action, and causal outcome}. This process yields a total of 12,000 interconnected evaluation samples. \rev{Although all clips share the same HRE framework, the questions are generated independently from the content of each clip rather than simply repeated across variants.} This design effectively tests the models' ability to capture microscopic cues for event reasoning at resolutions up to 4K.

\begin{figure}
  \centering
  % 第一张子图 (a)
  \begin{minipage}{0.35\textwidth}
    \centering
    \includegraphics[width=\linewidth]{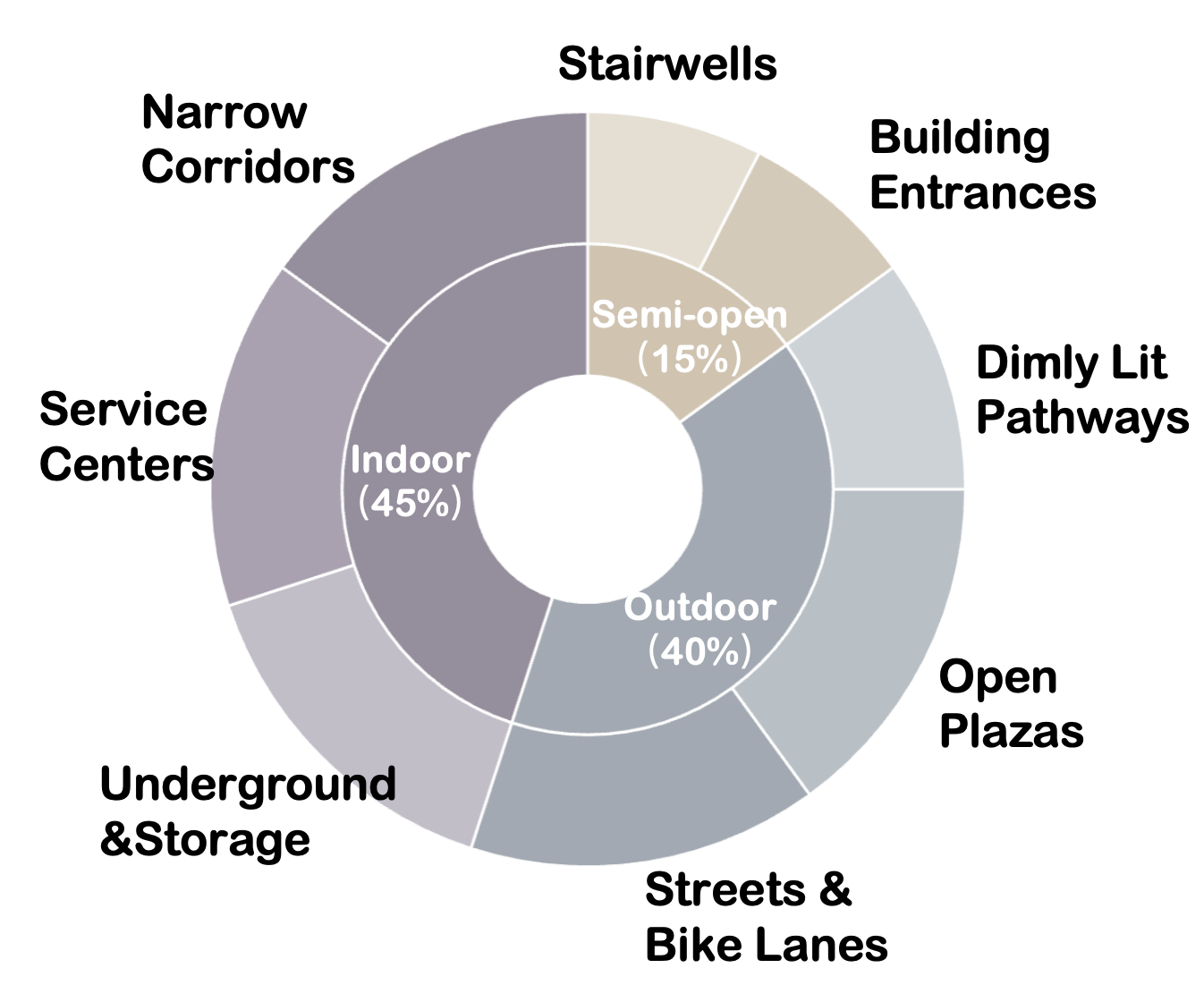}
    \centerline{(a)}
  \end{minipage}
  \hfill
  % 第二张子图 (b)
  \begin{minipage}{0.55\textwidth}
    \centering
    \includegraphics[width=\linewidth]{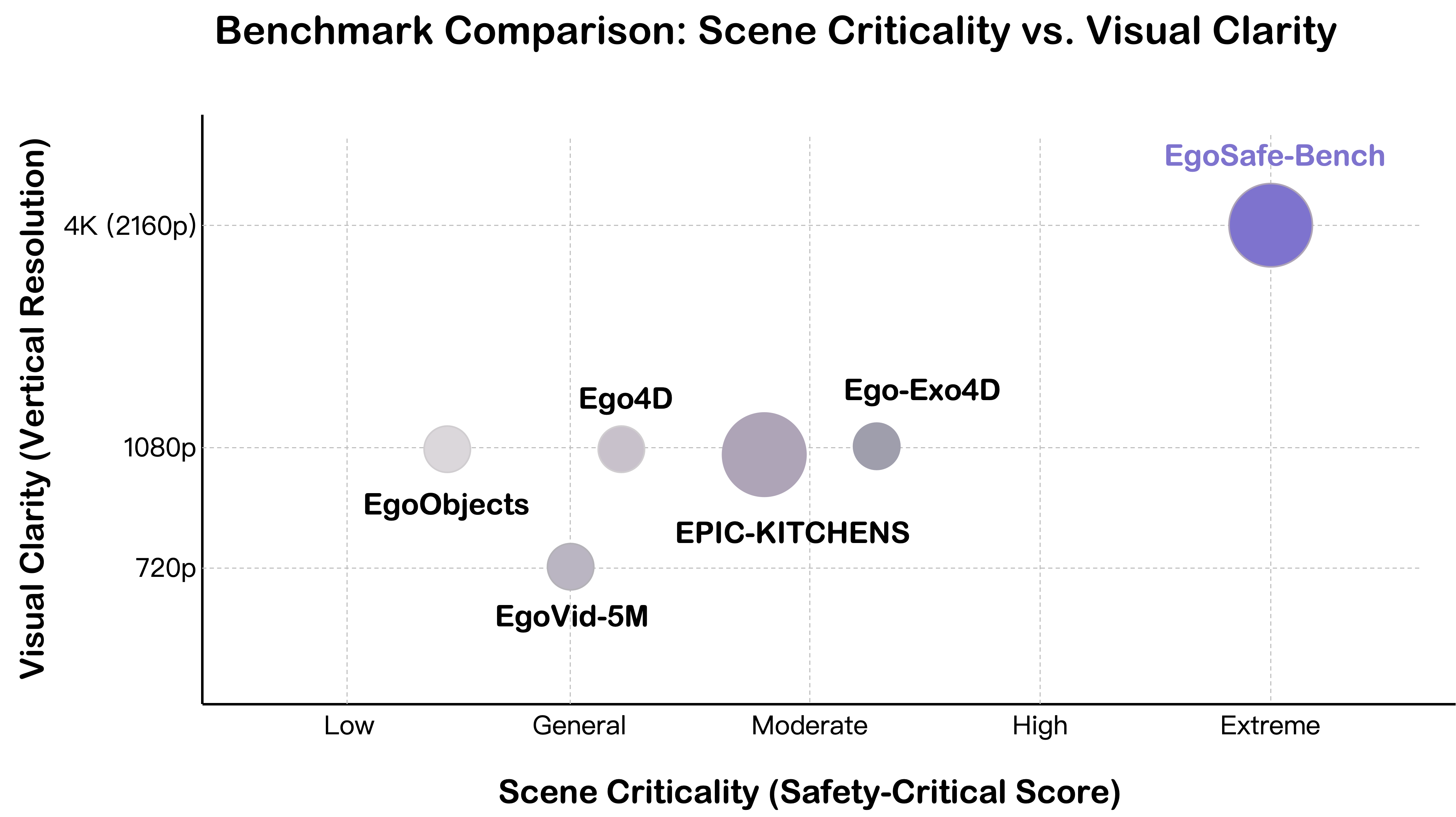}
    \centerline{(b)}
  \end{minipage}

  \caption{\textbf{Comprehensive analysis of scene diversity.} (b) Comparison of egocentric benchmarks where bubble size denotes annotation density.}
  \label{fig_comparison_combined}
\end{figure}

 Beyond logical depth, maintaining diverse environmental distribution~\cite{kollias2025dvd} is equally crucial. As illustrated by the sunburst chart in Figure~\ref{fig_comparison_combined}, we present the hierarchical breakdown of our recording environments. The inner ring categorizes the scenes into \textbf{indoor (45\%), outdoor (40\%), and semi-open (15\%) settings}. This proportional distribution is designed to closely mirror the statistical reality of everyday human activity spaces. Meanwhile, the outer ring details specific high-risk locations, such as \textit{``Narrow Corridors,'' ``Dimly Lit Pathways,'' and ``Building Entrances.''} This broad and quantitative distribution ensures our evaluation covers a wide range of practical applications, preventing models from overfitting to specific domain layouts.

This strategic emphasis on scene diversity is further justified when examining the landscape of existing egocentric benchmarks. As summarized in \textbf{Table~\ref{tab:egocentric_comparison}}, prominent datasets such as \textbf{Ego4D~\cite{Grauman_2022_CVPR}} and \textbf{EPIC-KITCHENS~\cite{9084270}} primarily facilitate research in \textit{daily activity recognition and skilled object manipulation}. While these benchmarks have advanced our understanding of human-object interactions, they are largely recorded \textbf{in controlled or soft environments}. Such settings often fail to capture the visual volatility and environmental stressors inherent in safety-critical encounters.

\begin{table*}
\centering
\caption{\textbf{Comparison with Mainstream Egocentric Video Datasets.} While existing large-scale egocentric datasets focus heavily on daily routines and explicit action recognition, EgoSafe-Bench pioneers the integration of evidence-based causal reasoning and systematic robustness testing in safety-critical scenarios.}
\label{tab:egocentric_comparison}
\resizebox{\textwidth}{!}{
\begin{tabular}{lcccccc}
\toprule
\textbf{Dataset} & \textbf{Core Domain} & \textbf{Task Focus} & \textbf{Annotation Type} & \textbf{NExT-QA-aligned Depth}\cite{xiao2021next,li2022from} & \textbf{Robustness Test} \\
\midrule
EPIC-Kitchens~\cite{9084270} & Cooking \& Kitchen & Action Recognition & Captions \& BBoxes & Descriptive & No \\
EgoObjects~\cite{zhu2023egoobjects} & Daily Objects & Object Detection & Bounding Boxes & Descriptive & No \\
EgoVid-5M~\cite{wang2024egovid} & General Vlogs & Video-Text Alignment & Text Captions & Descriptive & No \\
Ego4D~\cite{Grauman_2022_CVPR} & Daily Activities & Episodic Memory & Captions \& Timestamps & Temporal & No \\
Ego-Exo4D~\cite{grauman2024ego} & Skilled Activities & Skill Assessment & Trajectories \& Text & Temporal & No \\
\midrule
\rowcolor{gray!10} \textbf{EgoSafe-Bench (Ours)} & \textbf{Safety \& Security} & \textbf{Evidence-based Reasoning} & \textbf{4-Level Logic QA} & \textbf{Causal} & \textbf{Yes (11 Types)} \\
\bottomrule
\end{tabular}
}
\end{table*}

In addition to situational differences, a major gap exists in perceptual robustness. Most current first-person benchmarks~\cite{wang2024egovid,grauman2024ego,zhu2023egoobjects,Grauman_2022_CVPR,9084270} are captured in \textit{stable and clear conditions}, which may hide potential model weaknesses when facing real-world visual interference. By contrast, EgoSafe-Bench introduces a rigorous robustness evaluation~\cite{DBLP:conf/iclr/HendrycksD19} consisting of eleven types of environmental perturbations. This setup ensures that subtle visual signs remain detectable even under extreme camera shaking or poor lighting. Such \textit{a combination of high-risk scenarios and diverse visual challenges}~\cite{kollias2025dvd} fills a crucial void in egocentric vision, providing a more demanding testing ground for the reliability of large multimodal models.

\begin{table}[H]
\centering
\caption{\textbf{Comparison with State-of-the-Art Video Understanding Benchmarks.} EgoSafe-Bench fills a critical gap by combining First-Person perspective, high-resolution footage, and deep evidence-based reasoning.}
\label{tab:comparison}
\resizebox{\textwidth}{!}{
\begin{tabular}{lcccccc}
\toprule
\textbf{Dataset} & \textbf{Perspective} & \textbf{Resolution} & \textbf{Task Focus} & \textbf{Annotation Type} & \textbf{NExT-QA-aligned Depth}\cite{xiao2021next,li2022from} & \textbf{Robustness Test} \\
\midrule
UCF-Crime~\cite{Sultani_2018_CVPR} & 3rd-Person (CCTV) & Low & Anomaly Det. & Video Label & Descriptive & No \\
RWF-2000~\cite{9412502} & 3rd-Person & Low & Violence Det. & Video Label & Descriptive & No \\
XD-Violence~\cite{Wu2020not} & 3rd-Person & Mixed & Multimodal Det. & Audio-Visual & Descriptive & No \\
VVR (2024)~\cite{xiang2024video} & Mixed & Mixed & Violence Rating & Multimodal Rating & Descriptive & No \\
DVD (2025)~\cite{kollias2025dvd} & 3rd-Person & Mixed & Violence Det. & Video Label & Descriptive & No \\
\midrule
\rowcolor{gray!10}\textbf{EgoSafe-Bench (Ours)} & \textbf{1st-Person} & \textbf{Up to 4K} & \textbf{Evidence-based Reasoning} & \textbf{4-Level Logic QA} & \textbf{Causal} & \textbf{Yes (11 Types)} \\
\bottomrule
\end{tabular}
}
\end{table}

\begin{figure}[H]
  \centering
  % 第一张子图 (a)
  \begin{minipage}{0.40\textwidth}
    \centering
    \includegraphics[width=\linewidth]{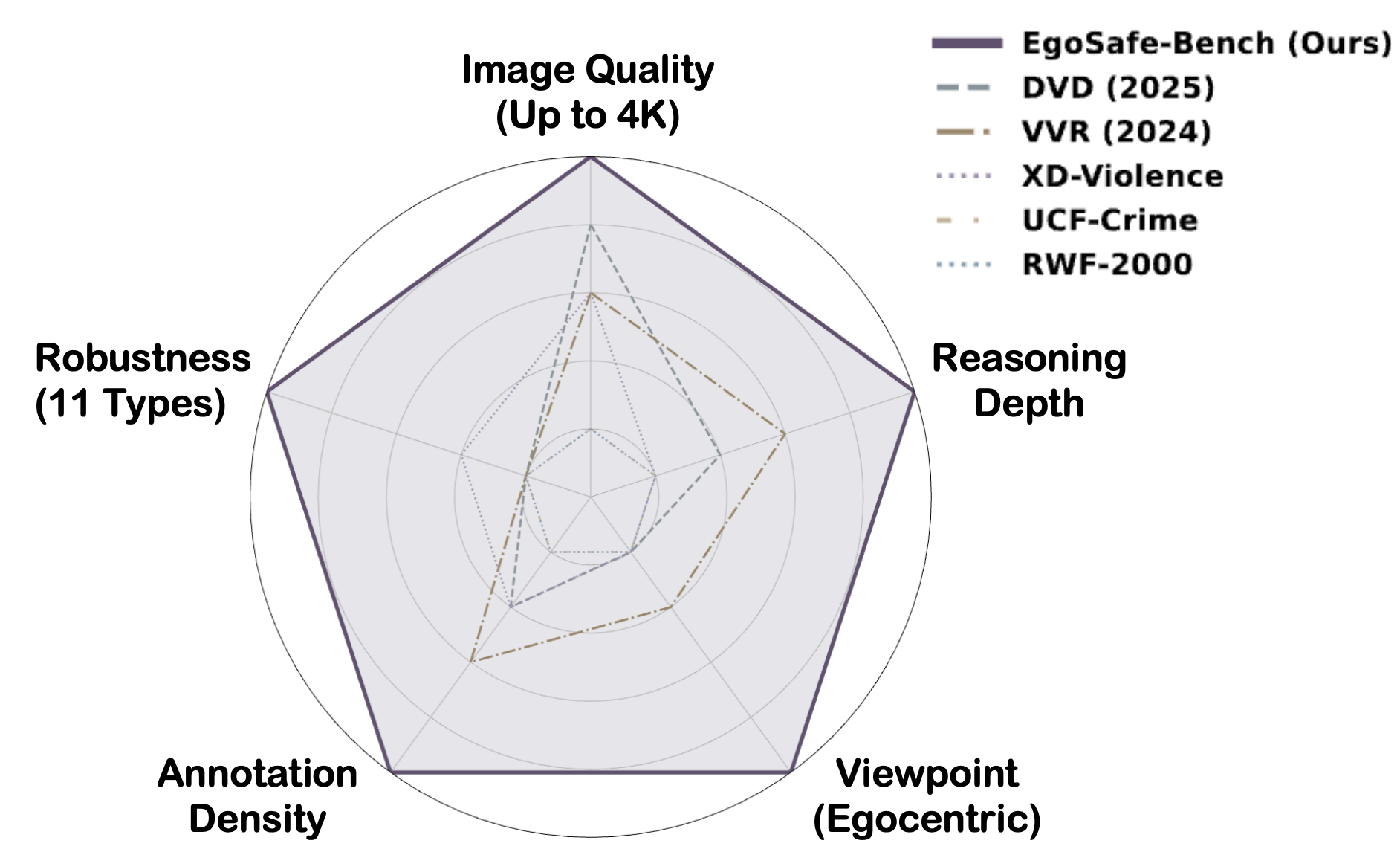}
    \centerline{(a)}
  \end{minipage}
  \hfill
  % 第二张子图 (b)
  \begin{minipage}{0.58\textwidth}
    \centering
    \includegraphics[width=\linewidth]{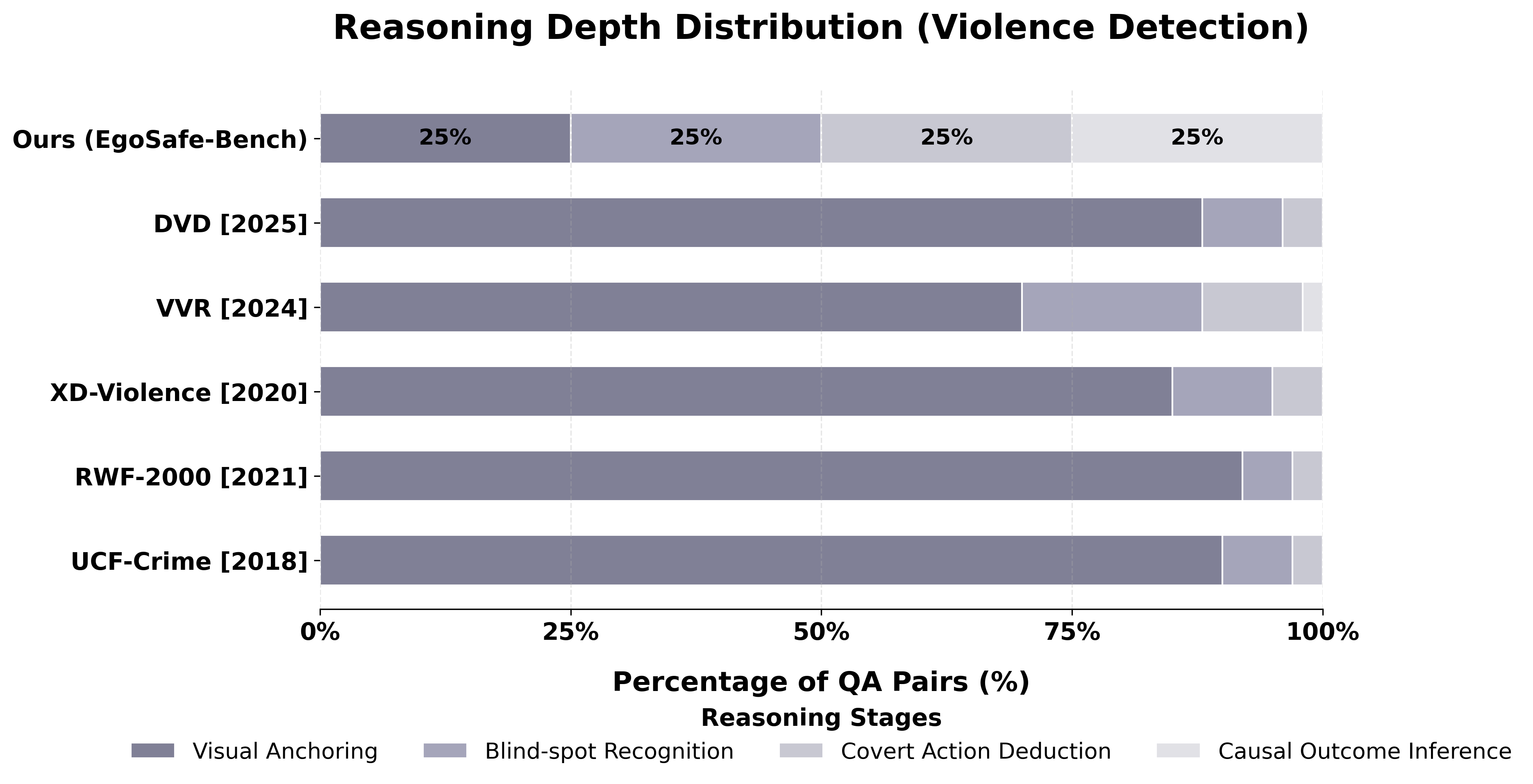}
    \centerline{(b)}
  \end{minipage}
  \caption{In-depth comparison of dataset capabilities and reasoning dimensions. (a) The radar chart evaluates benchmarks across five key metrics, illustrating that EgoSafe-Bench provides a superior balance of \textbf{image quality, robustness, and reasoning depth} compared to traditional violence detection datasets. (b) Distribution of reasoning categories across violence detection benchmarks. While existing datasets predominantly focus on visual anchoring, EgoSafe-Bench achieves a balanced distribution across all four hierarchical reasoning stages, specifically filling the void in covert action deduction and causal outcome inference.}
  \label{fig_comparisoned}
\end{figure}

Beyond the enhancement of environmental complexity, the most significant contribution of EgoSafe-Bench lies in its structured evaluation of reasoning depth. As illustrated by the comparative analysis in Table~\ref{tab:comparison} and Figure~\ref{fig_comparisoned}(b), traditional violence detection benchmarks~\cite{Sultani_2018_CVPR,9412502,Wu2020not,kollias2025dvd} are predominantly limited to recognition-based tasks. Their annotation frameworks focus heavily on Visual Anchoring, with nearly 90\% of QA pairs dedicated to identifying explicit actions or objects in stable frames. This representational bias often allows models to achieve high performance through superficial pattern matching, effectively bypassing the need for a true understanding of the underlying safety-critical dynamics.\cite{xiao2021next,li2022from}

In contrast, EgoSafe-Bench addresses this logical deficiency by introducing a \textbf{balanced 4-level reasoning hierarchy}. By distributing evaluation tasks evenly across \textit{Blind-spot Recognition, Covert Action Deduction, and Causal Outcome Inference}, our benchmark forces models to move beyond simple ``what is happening'' descriptions. This rigorous design requires large multimodal models to synthesize subtle environmental cues and predict potential hazards that are not immediately visible. By filling this void in causal and counterfactual reasoning, EgoSafe-Bench establishes a more comprehensive paradigm for assessing whether a model can truly interpret the complex intentionality and risks inherent in first-person hazardous encounters.

\section{Experiments}
\label{sec:exp}

% -----------------------------------------------------------------------------
% 4.1 Setup
% -----------------------------------------------------------------------------
\subsection{Experimental Setup}
\textbf{Models.} \rev{We evaluate a spectrum of state-of-the-art LVLMs, including the closed-source GPT-5 as a representative proprietary model, high-performance open-source models such as \textbf{Qwen3-VL}~\cite{qwen3technicalreport} and \textbf{VideoLLaMA 3}~\cite{damonlpsg2025videollama3}, and efficient edge-deployment candidates such as \textbf{MiniCPM-V 4.5}~\cite{yu2025minicpmv45cookingefficient} and \textbf{InternVL 3.5}~\cite{wang2025internvl3_5}.}

\noindent\textbf{Evaluation Metrics.} All models are evaluated in a zero-shot setting to assess their innate generalization. To disentangle descriptive fluency from reasoning, we employ a multi-dimensional metric suite. \textbf{SemSim (Semantic Similarity)} and \textbf{KFC (Key Fact Coverage)} establish a baseline for descriptive capacity by measuring the semantic alignment with the ground truth and the recall of atomic visual facts, respectively. The \textbf{Reasoning Score} is a composite metric derived from our HRE framework, heavily weighting the causal consistency between \textit{Blind Spot Construction} (Level 2) and \textit{Causal Result} (Level 4). Finally, to detect ``hallucinatory overconfidence'' under partial observability, we calculate the \textbf{ECE (Expected Calibration Error)}:$$\text{ECE}=\sum_{m=1}^{M}\frac{|B_m|}{N}\left|\text{acc}(B_m)-\text{conf}(B_m)\right|$$where $B_m$ represents the $m$-th of $M$ equally spaced confidence bins and $N$ is the total sample size. To ensure fair comparison and avoid prompt-induced variance, we extract the exact confidence $\text{conf}(B_m)$ for open-weight models by applying a softmax function over the raw logits of valid candidate tokens, faithfully reflecting the model's true limited visibility.

\noindent\textbf{Automated Evaluation Framework (GPT-5 Judge).} To scale the evaluation of complex evidence-based reasoning, all text-based metrics (SemSim, KFC, Reasoning Score) are computed using \textbf{GPT-5}~\cite{achiam2023gpt} as a gold-standard judge. Unlike traditional n-gram metrics (e.g., BLEU, CIDEr)~\cite{papineni2002bleu,vedantam2015cider} that fail to capture hidden causality, GPT-5 utilizes the provided ground truth reasoning chain to explicitly penalize hallucinated details and reward correct inferences. This ensures a rigorous evaluation standard focused on the validity of underlying evidence-based reasoning rather than surface-level text matching.

% -----------------------------------------------------------------------------
% 4.2 Main Results
% -----------------------------------------------------------------------------
\subsection{The Perception-Reasoning Decoupling}
\label{subsec:main_results}

Table~\ref{tab:main_results} presents the comparative performance across all models, revealing a critical \textbf{decoupling between perception and reasoning}.

\textbf{High Descriptive Fluency, Fragile Reasoning.}
\rev{GPT-5 achieves the strongest model scores (SemSim 85.40, KFC 78.20, Reasoning 74.60, and ECE 28.50), but remains below the human baseline, particularly at the intermediate HRE levels.} State-of-the-art models exhibit strong visual grounding but weak evidence-based deduction. \rev{Notably, \textbf{InternVL 3.5} achieves the highest descriptive performance among the open-source models (SemSim \textbf{81.35}, KFC \textbf{66.29}), yet its Reasoning Score drops significantly to \textbf{52.92}.} High scores in semantic and key fact metrics do not necessarily prove understanding, but only the ability to identify specific patterns within explicit frames. This gap persists across frontier models: \textbf{Qwen3-VL} achieves a commanding SemSim of \textbf{70.43} but falls sharply to \textbf{40.25} \rev{on the full benchmark} in reasoning, indicating a struggle to weave pixel-level observations into coherent narratives under occlusion. Similarly, \textbf{MiniCPM-V 4.5} maintains respectable SemSim (56.44) with poor reasoning consistency (28.96), while \textbf{VideoLLaMA 3} and \textbf{LLaVA-Video} plummet to \textbf{25.06} and \textbf{22.43}. Clearly, descriptive capacity does not equate to safety understanding.

\textbf{Human Performance Ceiling.}
\rev{We recruited 15 STEM-background volunteers completely independent of our data team.} \rev{Under a blind-testing protocol, volunteers completed single-choice assessments on all 12,000 samples after watching each associated video only once.} Each distractor was designed to represent a specific failure mode associated with one evaluation dimension; selecting that distractor was therefore counted as an error for the corresponding metric. \rev{Humans achieve highly consistent scores (SemSim 88.71, KFC 88.04, Reasoning 89.93, and ECE 19.92) with zero perception-reasoning decoupling.} \rev{The reasoning gap between GPT-5 (74.60) and the human baseline (89.93), together with the larger gap for the best open-source model (52.92), highlights a distinct cognitive deficit in current AI systems, confirming that EgoSafe-Bench's blind-spot inferences are deterministically solvable from context.}

% [TABLE 2: Main Results - Content Unchanged]
\begin{table}[t]
\centering
\caption{\textbf{Main Evaluation on EgoSafe-Bench.} We observe a distinct ``reasoning gap'': while Semantic Similarity (SemSim) remains high across top-tier models, the Reasoning Score (reflecting causal consistency in HRE) drops significantly. The \textbf{Human Baseline} demonstrates that the task is solvable, ruling out data ambiguity.}
\label{tab:main_results}
\resizebox{0.75\linewidth}{!}{%
\begin{tabular}{l|c|cc|cc}
\toprule
\textbf{Model} & \textbf{Param} & \textbf{SemSim}  & \textbf{KFC}  & \textbf{Reasoning Score}  & \textbf{ECE}   \\
\midrule
VideoLLaMA 3~\cite{damonlpsg2025videollama3} & 7B &50.17  &32.63  &25.06 &60.81   \\
InternVL 3.5~\cite{wang2025internvl3_5} & 8B & 81.35 & 66.29 & 52.92 & 73.48  \\
MiniCPM-V 4.5~\cite{yu2025minicpmv45cookingefficient} & 8B &56.44 &56.78  &28.96  &92.04   \\
Qwen3-VL~\cite{qwen3technicalreport} & 8B & 70.43 & 60.02 & 40.25 & 84.10  \\
LLava-Video~\cite{zhang2024llavanextvideo}{} & 7B &52.44 &42.85  &22.43  &48.46    \\
\midrule
\textbf{GPT-5} & - & \textbf{85.40} & \textbf{78.20} & \textbf{74.60} & \textbf{28.50} \\
\midrule
\textbf{Human Baseline}& - & \textbf{88.71} & \textbf{88.04} & \textbf{89.93} & \textbf{19.92}  \\
\bottomrule
\end{tabular}%
}
\end{table}

% -----------------------------------------------------------------------------
% 4.3 Hierarchical Logical Analysis
% -----------------------------------------------------------------------------
\subsection{Event-Based Reasoning Analysis}
To probe \textit{why} reasoning fails, we analyze the performance degradation across the four levels of our HRE framework.

\begin{figure}[t]
  \centering
  % \fbox{\rule{0pt}{1.8in} \rule{0.9\linewidth}{0pt}}
  \includegraphics[width=1\textwidth]{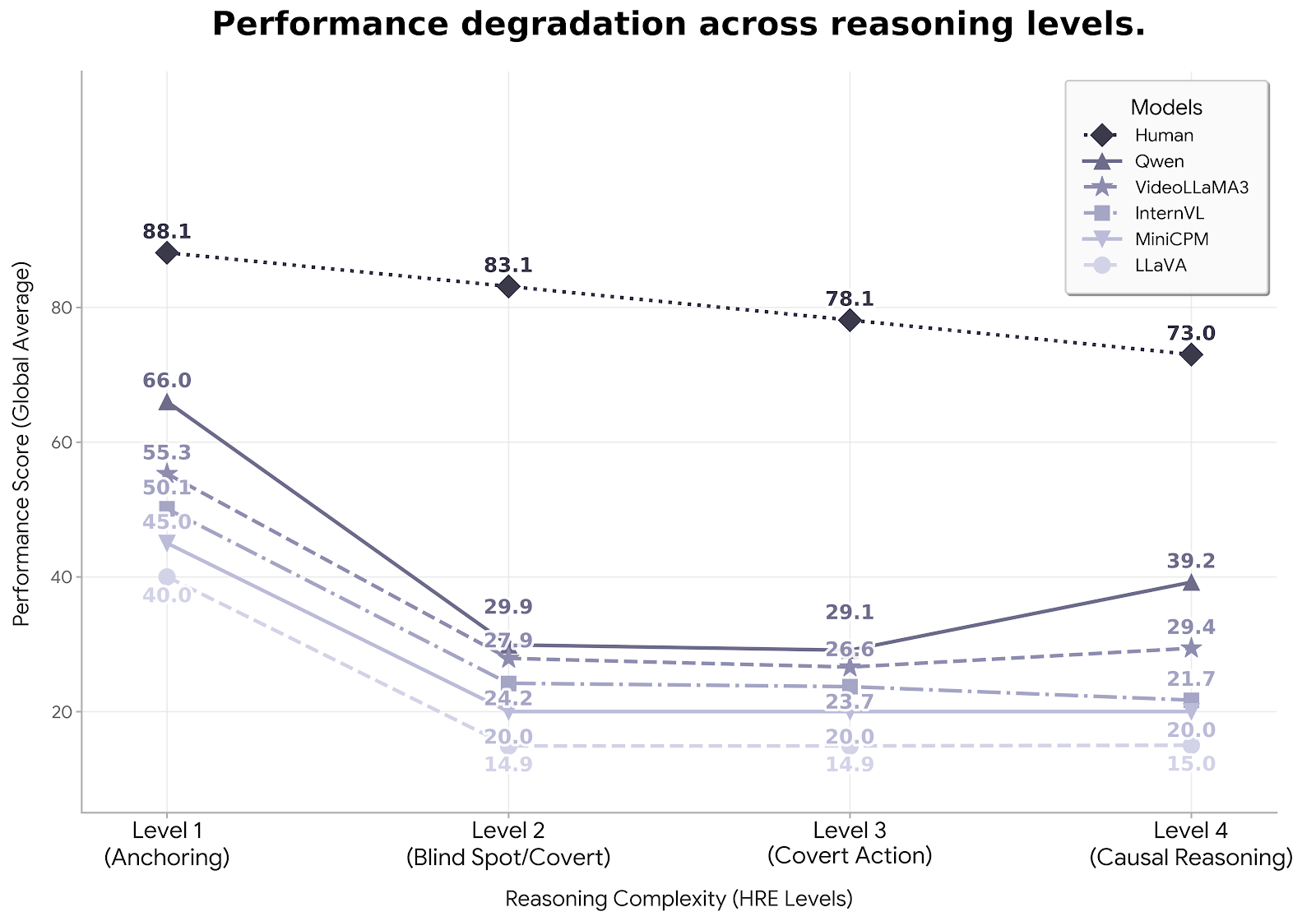}
  \caption{\textbf{Performance degradation across reasoning levels.} All models exhibit a downward trend as reasoning complexity increases. The steepest drop occurs at the Blind Spot/Covert Action stage.}
  \label{fig:hierarchical_drop}
\end{figure}

\textbf{The ``Blind Spot'' Bottleneck.} 
Performance remains stable at \textit{Level 1 (Anchoring)}, where visual evidence is explicit. However, we observe a sharp performance drop at \textit{Level 2 (Blind Spot Construction)} and \textit{Level 3 (Covert Action)}. \rev{As illustrated in Figure~\ref{fig:hierarchical_drop}, all open-source models shown exhibit a distinct downward trajectory as reasoning complexity increases.} The steepest drop occurs consistently at the Blind Spot/Covert Action stage. This suggests that current LVLMs struggle with \textbf{object persistence} in complex dynamic scenes. When an object enters a user-induced blind spot, models often fail to infer the invisible interaction based on surrounding cues, severing the reasoning chain required for safety assessment.

\begin{figure*}[t]
  \centering
  % \fbox{\rule{0pt}{2.5in} \rule{0.95\linewidth}{0pt}}
  \includegraphics[width=1\textwidth]{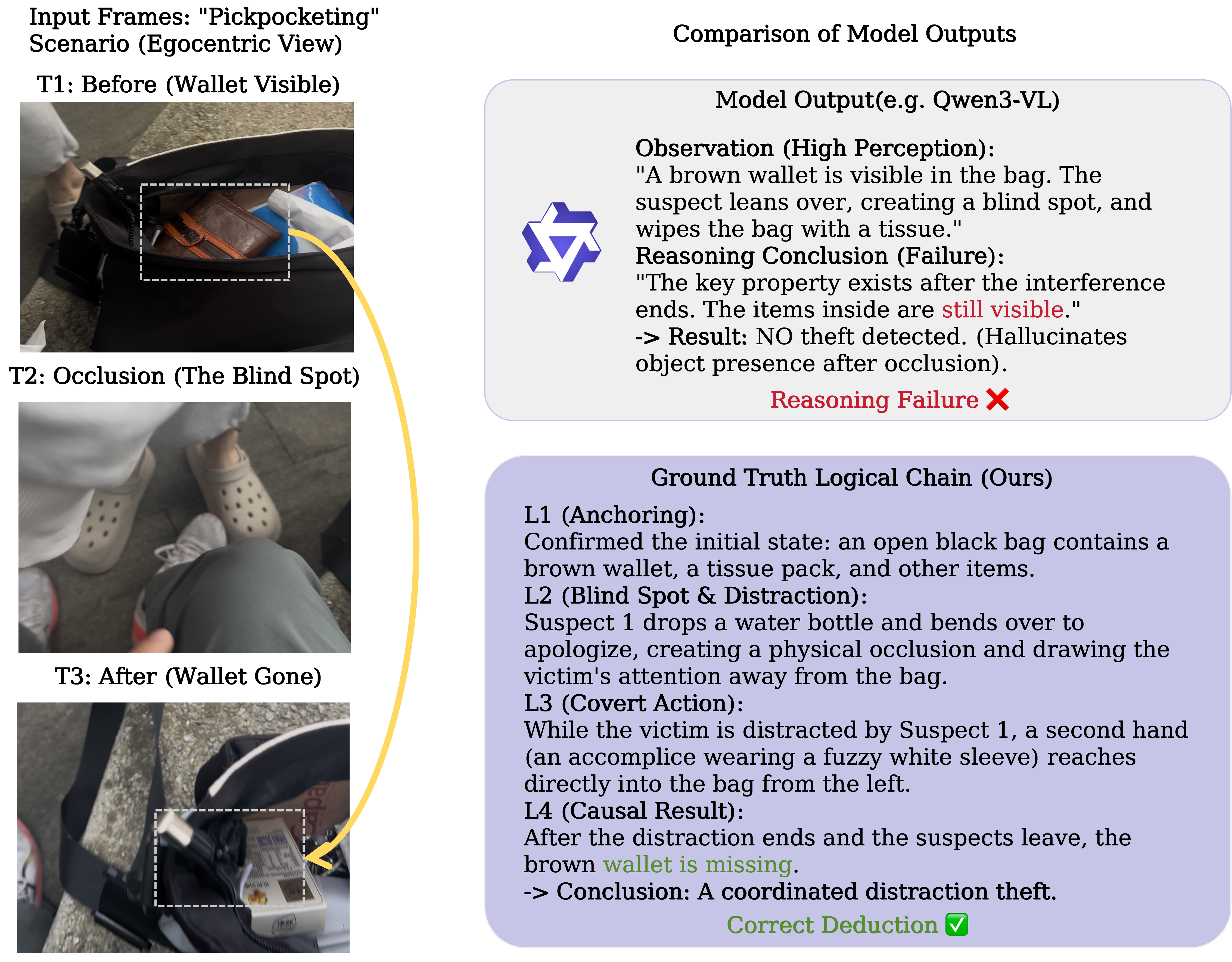}
  \caption{\textbf{Qualitative Analysis of Perception-Reasoning Decoupling.} Left: Input frames showing a ``Pickpocketing'' scenario where the wallet is occluded by a hand (Blind Spot). Right: Comparison of model outputs. While the model (e.g., Qwen3-VL) accurately describes the environment (\textcolor{green}{High Perception}), it hallucinates that the wallet is still visible or missing, failing to infer the theft (\textcolor{red}{Reasoning Failure}). In contrast, the Ground Truth Reasoning chain correctly deduces the covert action from the hand's motion.}
  \label{fig:qualitative_case}
\end{figure*}

\rev{This weakness is qualitatively demonstrated in our pickpocketing scenario (Figure~\ref{fig:qualitative_case}).} When the primary object (the wallet) is obscured by a physical occlusion (the blind spot), models like Qwen3-VL accurately describe the initial environment but frequently hallucinate the continued presence of the object after the occlusion. While models may achieve high scores in initial object identification, this does not necessarily prove genuine visual understanding, but only the ability to identify specific superficial patterns. Because they lack robust evidence-based deduction, they fail to utilize surrounding contextual clues---such as an accomplice's reaching motion---to deduce the covert action.

\rev{Interestingly, Figure~\ref{fig:hierarchical_drop} reveals a slight performance rebound at Level 4 (Causal Result) for select models.} This paradoxically underscores the necessity of the full HRE framework. Models can occasionally guess the correct final outcome without successfully navigating the intermediate logical steps, relying on shortcut predictions. \rev{By evaluating the entire QA chain, EgoSafe-Bench explicitly exposes and penalizes these shortcut-based predictions that bypass true causal deduction, ensuring the model maintains reasoning consistency.}

% [FIGURE 3: Unchanged]

% [FIGURE 4: Unchanged]

% -----------------------------------------------------------------------------
% 4.4 Robustness & Calibration
% -----------------------------------------------------------------------------
\subsection{Robustness Analysis}

\noindent\textbf{Impact of Egocentric Noise.} Unlike standard benchmarks, EgoSafe-Bench incorporates rigorous noise simulation (Table~\ref{tab:robustness_full}), revealing that noise disrupts temporal continuity for reasoning before affecting spatial features for recognition. Standard baselines exhibit severe fragility; notably, \textbf{LLaVA-Video} suffers a \textbf{44.4} average performance drop, highlighting the inadequacy of standard video instruction tuning for erratic egocentric dynamics. Even the most robust state-of-the-art model, \textbf{Qwen3-VL} (Reasoning Score 56.58 on clean data), experiences a substantial \textbf{31.2} decline. Specifically, \textit{Lighting} and \textit{Motion} degradations induce the most significant failures, suggesting that current top-tier models still rely heavily on clear visual cues rather than robust causal inference.

\noindent\textbf{Overconfidence under Partial Observability.} 
A critical safety requirement is that a model should express limited visibility when evidence is insufficient. However, our calibration analysis reveals a dangerous trend of \textbf{chronically overconfident hallucinations}.
\textbf{MiniCPM-V 4.5}, despite its efficiency, exhibits an alarming \textbf{ECE of 92.04} (Table~\ref{tab:main_results}). This implies that the model almost always answers with high confidence, even when hallucinating events in blind spots. 
\textbf{Qwen3-VL} also shows a high ECE (84.10), indicating that ``bigger'' models are not necessarily ``safer'' models regarding self-awareness. This high calibration error poses a significant risk for deploying current LVLMs in real-world safety monitoring, where a ``don't know'' response is preferable to a confident false positive.

% [TABLE 3: Robustness - Content Unchanged]
\begin{table}[h!]
\centering
\caption{\textbf{Comprehensive Robustness Evaluation on EgoSafe-Bench.} \rev{We report the Reasoning Score for the eight degradation conditions for which per-condition scores are available.} \rev{The complete generation protocol contains 11 controlled variants (Appendix~\ref{app:protocol}); no scores are inferred for the three unreported lighting conditions.} \textit{Avg. Drop} indicates the percentage performance decline compared to the Clean baseline.}
\label{tab:robustness_full}
\resizebox{\textwidth}{!}{%
\begin{tabular}{ll|ccccc} 
\toprule
\textbf{Noise Category} & \textbf{Specific Condition} & \textbf{VideoLLaMA 3} & \textbf{InternVL 3.5} & \textbf{Qwen3-VL} & \textbf{MiniCPM-V 4.5} & \textbf{LLaVA-Video} \\
 & & \textit{(7B)} & \textit{(8B)} & \textit{(8B)} & \textit{(8B)} & \textit{(7B)} \\
\midrule
\rowcolor{gray!10} \textbf{Baseline} & Clean (4K Raw) & 38.12 & 68.00 & 56.58 & 43.19 & 39.26 \\
\midrule
\multirow{2}{*}{\textbf{Resolution}} & 1080p Downsample & 25.74 & 56.00 & 44.58 & 27.65 & 26.84 \\
 & 720p Downsample & 23.85 & 55.00 & 42.58 & 27.97 & 25.16 \\
\midrule
\multirow{3}{*}{\textbf{Lighting}} & High Exposure (Mid) & 22.82 & 51.00 & 41.58 & 27.59 & 24.19 \\
 & High Exposure (Extreme) & 19.34 & 50.00 & 34.58 & 27.91 & 19.36 \\
 & Low Light (Deep) & 24.67 & 49.00 & 32.58 & 28.04 & 13.95 \\
\midrule
\multirow{2}{*}{\textbf{Motion}} & Gaussian Blur & 24.12 & 52.00 & 38.58 & 28.35 & 22.51 \\
 & Camera Shake & 22.31 & 48.00 & 36.58 & 26.93 & 18.93 \\
\midrule
\textbf{Geometry} & Horizontal Flip & 25.19 & 53.00 & 40.58 & 27.46 & 23.62 \\
\midrule
\multicolumn{2}{r|}{\textbf{\textit{Avg. Drop across all noise}}} & \textcolor{red}{-38.3\%} & \textcolor{red}{-23.9\%} & \textcolor{red}{-31.2\%} & \textcolor{red}{-35.8\%} & \textcolor{red}{-44.4\%} \\
\bottomrule
\end{tabular}
}
\end{table}

\subsection{Insufficiency of Existing Training Paradigms}
The reasoning failures of state-of-the-art LVLMs must be contextualized against their training backgrounds. The models we evaluated are pre-trained on massive corpora already encompassing mainstream egocentric (e.g., Ego4D~\cite{Grauman_2022_CVPR}) and causal video QA (e.g., NExT-QA~\cite{xiao2021next}) datasets. For instance, while frontier models like VideoLLaMA3~\cite{damonlpsg2025videollama3}  have already pretrained on Ego4D\cite{Grauman_2022_CVPR} dataset, achieve remarkable success on standard leaderboards like Next-QA~\cite{xiao2021next}, their severe performance degradation on EgoSafe-Bench reveals a fundamental limitation. This demonstrates that ordinary causal reasoning and benign first-person perception cannot trivially generalize to real-world safety scenarios. The unique combination of dynamic blind spots, covert actions, and severe physical noise makes EgoSafe-Bench a significantly more demanding cognitive challenge.

\section{Conclusion}
\label{sec:concl}
In this paper, we introduced EgoSafe-Bench, a pioneering benchmark specifically designed to evaluate evidence-based reasoning and causal deduction in first-person visual safety scenarios. To address the limitations of existing recognition-based datasets, EgoSafe-Bench comprises \textbf{3,000} high-fidelity mobile-captured video clips paired with \textbf{12,000} unique evaluation samples. These samples are strictly governed by our novel Hierarchical Reasoning Evaluation \textbf{(HRE)} framework, which mandates a four-tiered reasoning trajectory to ensure reasoning consistency and penalize shortcut predictions. Our comprehensive evaluations of state-of-the-art Large Vision-Language Models (LVLMs) expose a critical ``perception-reasoning decoupling''. While these frontier models exhibit strong semantic descriptive capabilities, their performance degrades sharply when tasked with multi-step causal deduction in dynamic blind spots. Furthermore, our robustness testing reveals that current models are highly susceptible to egocentric environmental noise and display dangerous overconfidence under limited visibility. By providing this rigorous, logic-driven testbed, EgoSafe-Bench aims to expose the cognitive gaps in current architectures and catalyze the development of genuinely trustworthy AI systems capable of reliable safety understanding in the real world.

\clearpage
\appendix
\renewcommand{\theHsection}{A.\arabic{section}}

\section{Dataset Construction and Controlled Degradation Protocol}
\label{app:protocol}
\rev{The 3,000 clips comprise 250 clean, staged source videos and 11 controlled variants per source.} \rev{All participants gave informed consent for academic use.} \rev{Before annotation and evaluation, faces and sensitive identifiers were irreversibly masked and manually double-checked.} \rev{The resulting de-identified dataset is intended for public academic research.}

\begin{table}[H]
\centering
\caption{\rev{Controlled variants generated for every clean source video.}}
\label{tab:degradation_protocol}
\small
\begin{tabular}{llc}
\toprule
\textbf{Category} & \textbf{Transformations} & \textbf{Count} \\
\midrule
Resolution & 1080p and 720p downsampling & 2 \\
Illumination & Three brighter and three darker settings & 6 \\
Motion & Gaussian blur and camera shake & 2 \\
Geometry & Horizontal flip & 1 \\
\midrule
\multicolumn{2}{r}{\textbf{Total controlled variants}} & \textbf{11} \\
\bottomrule
\end{tabular}
\end{table}

\section{Closed-Source Model Evaluation}
\label{app:gpt5}
\rev{Table~\ref{tab:gpt5_levels} compares GPT-5 with selected open-source baselines and the human reference.} \rev{GPT-5 approaches the human scores at Levels 1 and 4, but still trails on Levels 2 and 3, and the sharp Level 1 to Level 2 drop persists.} \rev{This pattern confirms the value of HRE as a diagnostic protocol for multi-stage safety reasoning.}

\begin{table}[H]
\centering
\caption{\rev{GPT-5 evaluation compared with selected models.} \rev{L1--L4 denote Anchoring, Blind Spot, Covert Action, and Causal Outcome.}}
\label{tab:gpt5_levels}
\small
\setlength{\tabcolsep}{3pt}
\begin{tabular}{lcccccccc}
\toprule
\textbf{Model} & \textbf{SemSim} & \textbf{KFC} & \textbf{Rsn} & \textbf{ECE} & \textbf{L1} & \textbf{L2} & \textbf{L3} & \textbf{L4} \\
\midrule
InternVL 3.5 & 81.4 & 66.3 & 52.9 & 73.5 & 50.1 & 24.2 & 23.7 & 21.7 \\
Qwen3-VL & 70.4 & 60.0 & 40.3 & 84.1 & 66.0 & 29.9 & 29.1 & 39.2 \\
\midrule
\textbf{GPT-5} & \textbf{85.4} & \textbf{78.2} & \textbf{74.6} & \textbf{28.5} & \textbf{82.3} & \textbf{68.5} & \textbf{65.2} & \textbf{71.4} \\
Human & 88.7 & 88.0 & 89.9 & 19.9 & 88.1 & 83.1 & 78.1 & 73.0 \\
\bottomrule
\end{tabular}
\end{table}

\section{Single-Level HRE Ablation}
\label{app:ablation}
\rev{To verify the necessity of HRE, we remove one level at a time and require the model to transition directly between the remaining stages.} \rev{We report Qwen3-VL 8B results scored with GPT-5 as the judge.} \rev{Removing any single level causes an average drop of 11.62 points in the Reasoning Score, with the largest drop when Level 2 is removed.} \rev{The ablation was conducted only on Qwen3-VL; no results are inferred for the remaining models.}

\begin{table}[H]
\centering
\caption{\rev{Single-level HRE ablation on Qwen3-VL 8B.}}
\label{tab:hre_ablation}
\small
\setlength{\tabcolsep}{3pt}
\begin{tabular}{lc}
\toprule
\textbf{HRE configuration} & \textbf{Reasoning Score} \\
\midrule
Full HRE (L1 $\rightarrow$ L2 $\rightarrow$ L3 $\rightarrow$ L4) & \textbf{40.25} \\
\midrule
w/o L1 (L2 $\rightarrow$ L3 $\rightarrow$ L4) & 32.14 \\
w/o L2 (L1 $\rightarrow$ L3 $\rightarrow$ L4) & 25.30 \\
w/o L3 (L1 $\rightarrow$ L2 $\rightarrow$ L4) & 28.45 \\
\bottomrule
\end{tabular}
\end{table}

\bibliographystyle{splncs04}
\bibliography{main}
\end{document}